\documentclass{ieeeaccess}
\usepackage{cite}
\usepackage{amsmath,amssymb,amsfonts}
\usepackage{algorithmic}
\usepackage{graphicx}
\usepackage{textcomp}
\usepackage[pagebackref=true,breaklinks=true,colorlinks,bookmarks=false]{hyperref}

\usepackage{bm}
\makeatletter
\AtBeginDocument{\DeclareMathVersion{bold}
\SetSymbolFont{operators}{bold}{T1}{times}{b}{n}
\SetSymbolFont{NewLetters}{bold}{T1}{times}{b}{it}
\SetMathAlphabet{\mathrm}{bold}{T1}{times}{b}{n}
\SetMathAlphabet{\mathit}{bold}{T1}{times}{b}{it}
\SetMathAlphabet{\mathbf}{bold}{T1}{times}{b}{n}
\SetMathAlphabet{\mathtt}{bold}{OT1}{pcr}{b}{n}
\SetSymbolFont{symbols}{bold}{OMS}{cmsy}{b}{n}
\renewcommand\boldmath{\@nomath\boldmath\mathversion{bold}}}
\makeatother

\def\BibTeX{{\rm B\kern-.05em{\sc i\kern-.025em b}\kern-.08em
    T\kern-.1667em\lower.7ex\hbox{E}\kern-.125emX}}

\newcommand{\etal}{\textit{et~al.}}
\newcommand{\eg}{e.g.,}
\newcommand{\ie}{i.e.,}

\begin{document}
\history{Date of publication xxxx 00, 0000, date of current version xxxx 00, 0000.}
\doi{10.1109/ACCESS.2024.0429000}

\title{ShapeGraFormer: GraFormer-Based Network for Hand-Object Reconstruction from a Single Depth Map}
\author{\uppercase{Ahmed Tawfik Aboukhadra}\authorrefmark{1, 2}, 
\uppercase{Jameel Malik}\authorrefmark{3}, 
\uppercase{Nadia Robertini}\authorrefmark{1}, 
\uppercase{Ahmed Elhayek}\authorrefmark{4}, and 
\uppercase{Didier Stricker}\authorrefmark{1, 2}.
}

\address[1]{Augmented Vision Group, German Research Center for Artificial Intelligence (DFKI), 67663 Kaiserslautern, Germany}
\address[2]{Department of Computer Science, University of Kaiserslautern-Landau (RPTU), 67663 Kaiserslautern, Germany}
\address[3]{School of Electrical Engineering and Computer Science (SEECS), National University of Sciences and Technology (NUST), 44000 Islamabad, Pakistan}
\address[4]{Artificial Intelligence Department,
College of Computer and Cyber Sciences,
University of Prince Mugrin (UPM), 42241 Madinah, Saudi Arabia}

\tfootnote{This work was partially funded by the Federal Ministry
of Education and Research of the Federal Republic of Germany (BMBF),
under grant agreements: DECODE [Grant Nr 01IW21001], and GreifbAR [Grant
Nr 16SV8732].}

\markboth
{Aboukhadra \headeretal: ShapeGraFormer: GraFormer-Based Network for Hand-Object Reconstruction from a Single Depth Map}
{Aboukhadra \headeretal: ShapeGraFormer: GraFormer-Based Network for Hand-Object Reconstruction from a Single Depth Map}

\corresp{Corresponding author: Ahmed Tawfik Aboukhadra (e-mail: ahmed\_tawfik.aboukhadra@dfki.de).}

\begin{abstract}
3D reconstruction of hand-object manipulations is important for emulating human actions.
Most methods dealing with challenging object manipulation scenarios focus on hands reconstruction in isolation, ignoring physical and kinematic constraints due to object contact.
Some approaches produce more realistic results by jointly reconstructing 3D hand-object interactions.
However, they focus on coarse pose estimation or rely upon known hand and object shapes.
We propose an approach for realistic 3D hand-object shape and pose reconstruction from a single depth map.
Unlike previous work, our voxel-based reconstruction network regresses the vertex coordinates of a hand and an object and reconstructs more realistic interaction. 
Our pipeline additionally predicts voxelized hand-object shapes, having a one-to-one mapping to the input voxelized depth. 
Thereafter, we exploit the graph nature of the hand and object shapes, by utilizing the recent GraFormer network with positional embedding to reconstruct shapes from template meshes.
In addition, we show the impact of adding another GraFormer component that refines the reconstructed shapes based on the hand-object interactions and its ability to reconstruct more accurate object shapes. From those contributions, we name our system $ShapeGraFormer$.
We perform an extensive evaluation on the HO-3D and DexYCB datasets and show that our method outperforms existing approaches in hand reconstruction and produces plausible reconstructions for the objects. 

\end{abstract}


\titlepgskip=-21pt

\maketitle

\section{Introduction}
\label{sec:intro}

Understanding and reconstructing hand and object interactions in 3D is important for analyzing and imitating human behavior. 
Modeling hand-object interactions realistically has applications in several fields, including robotics, virtual reality, and augmented reality.
The last decade has witnessed rapid advance in 3D hand pose~\cite{malik2018deephps,malik2018structure,ge20193d,Kulon_2020_CVPR,zhang2019end,yang2020seqhand,malik2019whsp,Zhou_2020_CVPR,HandVoxNet2020} and object ~\cite{Tzionas2015,Tekin2018,Wang2019,Peng2019,Wang2020} estimation in isolation.
In contrast, reconstructing a hand and an object simultaneously from a monocular image has received lesser attention. 
Besides the common issues from complex pose variation, clutter, and self-occlusion, methods for reconstructing hands and objects that are in close contact have to cope with \textit{mutual occlusions}. 
Existing methods, dealing with challenging object manipulation scenarios, tend to focus on hand reconstruction alone~\cite{Mueller2017RealTimeHT,baek2020weakly}. Recent approaches to jointly reconstruct hand and object, often neglect the intrinsic kinematic and physical correlation, which exists among the two~\cite{sridhar2016real,mueller2018ganerated,hopenet,Aboukhadra_2023_WACV}.
Approaches exploiting that mutual relation typically focus on coarse pose estimation or assume known hand and object shapes~\cite{Tzionas2016,Tsoli2018,hasson2019learning,tekin2019h+,zhang2019interactionfusion,Zhe2020,hasson2020leveraging}.

In this paper, we propose one of the first approaches to jointly reconstruct physically valid hand and object shapes from a single depth map. In contrast to most methods, ours can generalize to different hand models and unknown object shapes, by directly regressing mesh vertices, rather than model parameters.
We avoid perspective distortion and scale ambiguities, typical for RGB image-based methods by working exclusively in the 3D domain.
The input of our deep network is a 3D voxelized grid of a given depth map, centered around the hand-object interaction. The output consists of (i) 3D heatmaps that describe the location of hand-object pose keypoints (ii) hand-object shape predictions in voxelized form, and (iii) the corresponding 3D hand-object mesh vertex coordinates.

To effectively tackle the problem of simultaneous hand-object pose and shape reconstruction, we propose a novel architecture based on a Graph Convolutional network and Multi-headed Attention layers. Specifically, we introduce the following novel modules:
\begin{enumerate}
    \item $PoseNet$ \& $VoxelNet$: Two 3D-to-3D voxel-based networks for hand-object pose and shape estimation, respectively;
    \item $ShapeGraFormer$: State-of-the-art GraFormer (Transformers with Graph Convolutional layers) for Hand-Object shape reconstruction;
    \item Positional Embedding layer based on the template meshes for the hand and the object;
    \item Topologically consistent object mesh registration for optimal object modeling and shape prediction;
\end{enumerate}

We validate our design choices and evaluate our approach both quantitatively and qualitatively. Our approach outperforms previous work on popular datasets~\cite{hampali2020honnotate,dexycb}, as also reported on the challenge website \footnote{\url{https://codalab.lisn.upsaclay.fr/competitions/4393\#results}}, with a minimal shape reconstruction improvement of $0.43$ cm over the state-of-the-art. 
\section{Related Work}
\label{sec:related_work}
%
%
In this section, we discuss the existing methods for joint hand-object reconstruction from challenging monocular object manipulation scenarios. For a survey of works focusing on the reconstruction of hands and objects in isolation, please refer respectively to~\cite{HandVoxNet2020} and~\cite{Tekin2018}.

Most methods that jointly reconstruct 3D hand-object from monocular take single RGB~\cite{tekin2019h+,hasson2019learning,Zhe2020,hasson2020leveraging,hampali2022keypoint,yang2022artiboost,Aboukhadra_2023_WACV} or RGB+D input~\cite{Tzionas2016,Tsoli2018,sridhar2016real}. Very few recent approaches consider single depth input~\cite{Oberweger2019,Chen2022,Zhang2021}, due to its intrinsic challenges. Nevertheless, the availability of depth information is a key factor in allowing proper, in-scale, and scene-dependent 3D shape reconstruction, required for \eg~virtual and augmented reality applications, especially for single-frame, one-shot approaches. While RGB-based approaches can rely on large labeled data for training their models ~\cite{garcia2018first,mueller2018ganerated,hasson2019learning}, typically based on the MANO parametric hand model (hand Model with Articulated and Non-rigid defOrmations)~\cite{romero2017embodied}, methods that rely on the depth channel have access to only a limited amount of data, because labeling of real scenes in the 3D domain are impractical. For this reason, most depth as well as some RGB+D approaches build their synthetic datasets in an attempt to improve their results~\cite{Oberweger2019,hasson2019learning,Chen2022,Zhang2021,sridhar2016real}. 
Recently, some datasets have been introduced to bridge the gap between RGB and depth data availability, \ie~HO3D and DexYCB~\cite{hampali2020honnotate,dexycb}. Still, they only provide limited sample variation, especially in terms of number of objects considered. As a result, modeling and reconstructing the object of interaction remains an underconstrained, challenging problem. Many methods restrict themselves to objects with known shapes~\cite{Tzionas2016,Tsoli2018} or reconstruct them on the fly, under certain object shape and visibility assumptions~\cite{Zhang2021,Groueix2018,Kato2018}. The remaining approaches only output coarse object pose, \eg~represented with bounding boxes~\cite{tekin2019h+,hampali2022keypoint,hopenet,yang2022artiboost}. Some more sophisticated methods, define a deformable object model, typically based on a 3D sphere, capable of coarsely adapting to virtually any convex shape~\cite{hasson2019learning,Groueix2018,Aboukhadra_2023_WACV}. The resulting object reconstruction typically lacks surface details, due to over-smoothing. Nevertheless, we believe it to be a strong base with sufficient surface information to reliably reconstruct interactions with the hand surface.
The remaining hand reconstruction as a standalone problem has been largely studied in the past~\cite{Huang2021}. However, modeling and reconstructing 3D hand-object interactions is still very challenging, especially in a monocular setting due to the large mutual occlusions. To simplify the task, many approaches focus on fewer model parameter estimation~\cite{hasson2019learning,Zhe2020}. In contrast, we directly regress hand-object vertices, which makes our method capable to generalize to different models or geometries.
While many approaches independently reconstruct hand and object before putting them in context~\cite{hasson2019learning,Zhe2020}, our approach is designed to simultaneously regress hand-object geometries, thus allowing us to implicitly study the underlying physical and kinematic correlation existing among the two.
Another advantage of direct vertex regression is the possibility of reconstructing extra, more realistic deformation, which cannot be synthesized via model-parameter tuning. In this case, to avoid distortions in the 3D reconstructions, though, particular care has to be given to the algorithm design, especially to the shape estimation components. To avoid perspective distortion from the start, we convert the input depth map to a 3D point cloud and base the remaining algorithmic steps on the corresponding voxelized domain, following the approach from Malik~\etal~\cite{HandVoxNet2020} (HandVoxNet). The core of our pipeline is based on a Graph Convolutional Network (GCN), which has been shown to effectively tackle shape reconstruction problems on graph-structured data, such as mesh topology~\cite{HandVoxNet++2021}.
In contrast to the RGB-based approach by Aboukhadra~\etal~\cite{Aboukhadra_2023_WACV} (THOR-Net), our depth-based method reconstructs hand-object geometries in one shot, thus significantly reducing computational costs and training time. We qualitatively and quantitatively demonstrate the effectiveness of our $ShapeGraFormer$ component, comprising a combination of GCN and Multi-headed Attention layers, as in~\cite{Graformer2022}, in the simultaneous reconstruction of hand and object interaction as well as in shape refinement. For the latter, we demonstrate its effectiveness in improving physical hand-object interactions, without the explicit need for expensive physical simulation~\cite{Tzionas2016} or penetration and contact loss, as required in previous work~\cite{Zhe2020,hasson2019learning}.
\section{Method}
\begin{figure*}[!t]
\begin{center}
  \includegraphics[width=\linewidth]{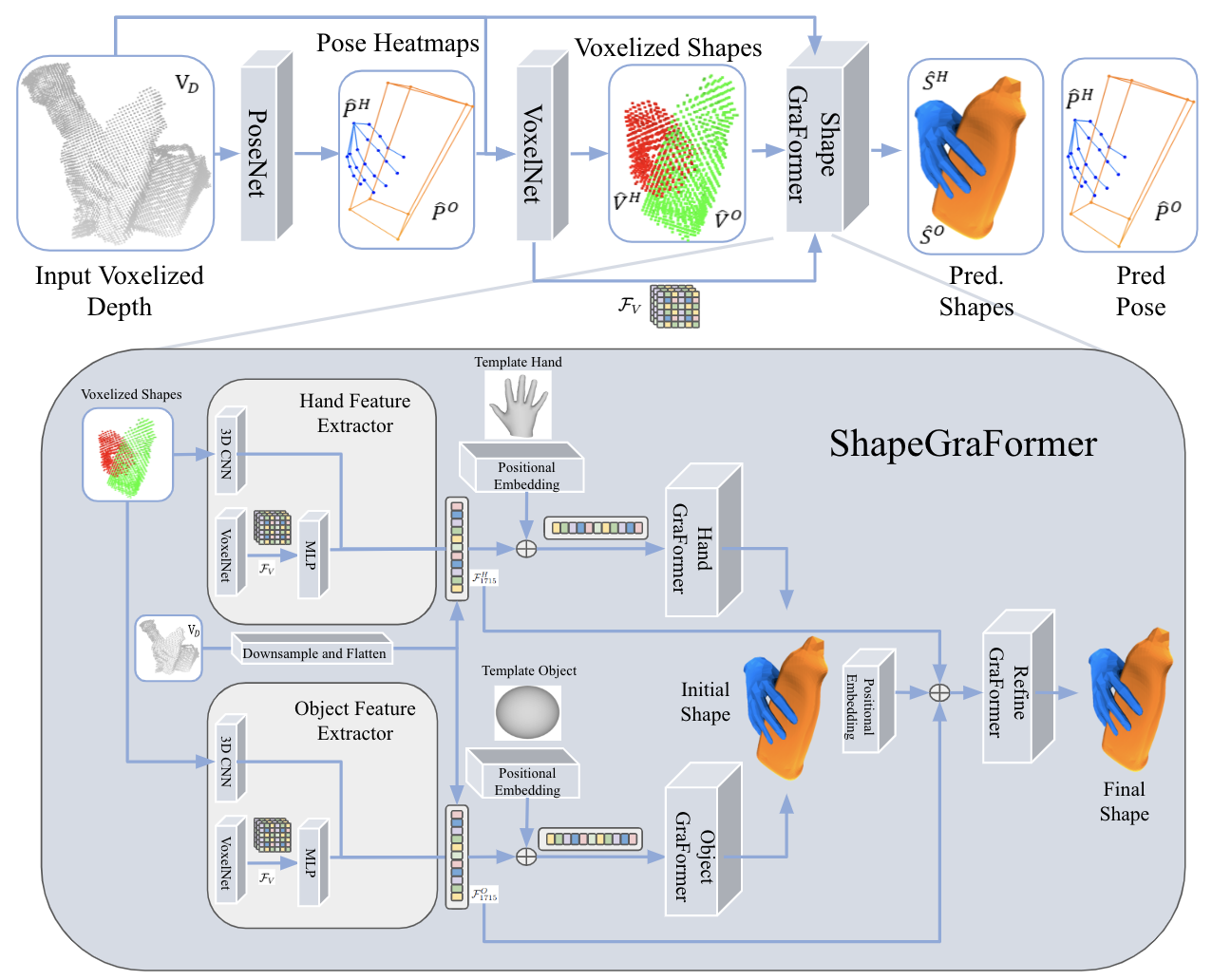}
\end{center}
  \caption{Overview of our pipeline. Our method takes as input a single depth image that is converted to a 3D voxelized representation and outputs 3D realistic hand-object interactions. The input depth is first forwarded through a sequence of three components, namely $PoseNet$, $VoxelNet$, and $ShapeGraFormer$, predicting respectively hand-object pose heatmaps, voxelized shape representations, and topologically-consistent shapes. 
  }
\label{fig:pipeline}
\end{figure*}

We design a voxel-based 3D CNN along with a $ShapeGraFormer$ network to reconstruct plausible 3D hand-object shapes in a single forward pass from an input depth image. 
Our pipeline
is depicted in Fig~\ref{fig:pipeline}.
In a preprocessing step, we convert the input depth map to its voxelized form $V_D$ by projecting the raw depth image pixels into a cubic binary 3D grid around the hand-object interaction space, similar to~\cite{HandVoxNet2020}. 
Given $V_D$, the first network component in the pipeline, $PoseNet$, predicts 3D hand and object pose in the form of 3D heatmaps, resp. $\hat{P}^H$ and $\hat{P}^O$, see Section~\ref{sec:posenet}. The resulting heatmaps concatenated with $V_D$ are forwarded to the second network component, $VoxelNet$, which produces a voxelized shape representation of the hand and the object, resp. $\hat{V}^H$ and $\hat{V}^O$, see Section~\ref{sec:voxelnet}.
The voxelized depth $V_D$ along with the intermediate voxelized representation and the features of the $VoxelNet$ serve as input to the next network component, $ShapeGraFormer$, which regresses topologically consistent hand and object vertices,
see Section~\ref{sec:graformer}. 
We describe our hand and object models in the next Section.

\subsection{Hand and Object Models}
\label{sec:model}
\paragraph{Hand Model}
To represent hands, we use the MANO parametric hand model~\cite{romero2017embodied}, which maps joint angles and shape parameters to a triangulated mesh representation $S^H:=\{v^H, f^H\}$, consisting of $|v_H|=778$ 3D vertices, and a set of 3D hand-skeleton joints $\mathcal{J}$, where $|\mathcal{J}|=21$, representing the hand pose. 
%

\paragraph{Object Model}
Since objects differ greatly in terms of shape and size, direct regression of object vertices from a set of topologically inconsistent meshes results in strong noise in the output. In order to bring all the known object geometries into the same topologically consistent representation $S^O :=\{v^O, f^O\}$, we deform a source mesh via a set of vertex displacements $D^j := \{d^j_i, \forall v^O_i\}$, scale $c_j$ and translation $t_j$ to approximate all target objects $S^j$ as:
\begin{equation}
    S^j \sim c^j (S^O + D^j) + t^j, \forall j
\end{equation}
Our source mesh is a sphere obtained performing $4$ subdivisions of a unit, \ie radius equals 1, regular icosahedron centered at the origin, each subdivision generating four new faces per face, resulting in a total of $|v^O| = 2562$ vertices and $|f^O| = 5120$ faces. The object pose is defined as the 3D bounding box, consisting of eight 3D corner coordinates.
%

\paragraph{Object Approximation Procedure}
To approximate differently shaped objects, we first scale ($c^j$) and translate ($t^j$) the target mesh $S^j$ to fit inside the sphere. Then, to learn the set of displacements $D^j$, we minimize the chamfer distance $E^{C}$ between the predicted and target mesh computed on a total of $5,000$ surface samples. We additionally enforce surface smoothness by adding the following shape regularizers to the objective: (i) surface laplacian smoothness $E^{L}$ which ensures smooth vertex positions between neighboring vertices, (ii) normals consistency $E^{N}$ across neighboring faces, and (iii) edge length consistency $E^{E}$ across the entire deformed mesh which minimizes the length of edges. We minimize the weighted summation of all mentioned terms in the following equation using stochastic gradient descent (SGD). The full implementation of spherical object approximation can be found in the PyTorch3D library\footnote{\url{https://pytorch3d.org/tutorials/deform_source_mesh_to_target_mesh}}.
\begin{equation}
   E = w_C E^{C} + w_L E^{L} + w_N E^{N} + w_E E^{E}
\end{equation}
%
\begin{figure}
\begin{center}
  \includegraphics[width=\linewidth]{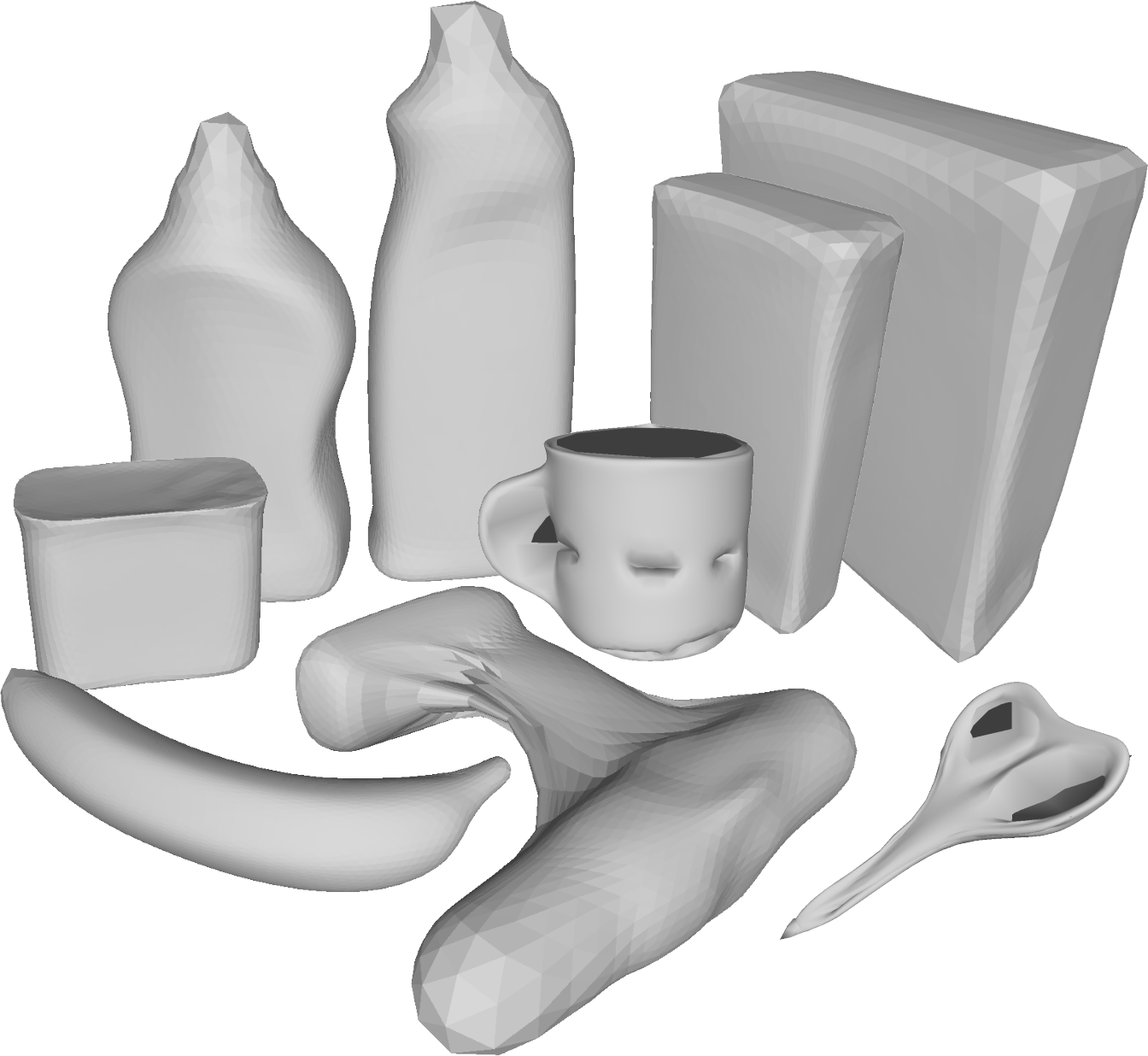}
\end{center}
  \caption{The complete collection of our sphere-based object approximations used as ground-truth.}
\label{fig:objects}
\end{figure}
\subsection{PoseNet: 3D Pose Estimation}
\label{sec:posenet}
The first component of our network pipeline, $PoseNet$, simultaneously estimates 3D hand joint location and 3D bounding box corners of the object from an input voxelized depth map.
We modify the V2V-PoseNet architecture of ~\cite{moon2017v2v} by introducing a new $1\times1\times1$ volumetric convolutional back-layer, targeted to predict 3D object pose in parallel to the 3D hand pose.
The input depth map is converted to a 3D binary voxelized representation $V_D$ of size $88\times88\times88$.  Each voxel value is set to $1$ when occupied and $0$ otherwise, \ie $V_D \in \{0,1\}$. The cubic size is fixed to the empirically found value of $200$ mm. The output of the network consists of a set of 3D heatmaps: (i) One set for each $j$ hand joints coordinates $\hat{P}^H_j(v)$, and (ii) one for each $b$ object bounding box corners coordinates $\hat{P}^O_b(v)$. All heatmaps are discretized on a grid of size $44\times44\times44$. 
Ground-truth heatmaps $P_j^H(v)$ and $P_b^O(v)$ are generated by applying a 3D Gaussian centered on the ground-truth locations with fixed standard deviation. The training loss between the predicted and targeted heatmaps is calculated by mean-squared error (MSE):
\begin{equation}
    \mathcal{L}_{pose} = \sum_v \sum_{j = 1}^{|J|} || \hat{P}^H_j - P^H_j || + \sum_{b = 1}^{|B|} || \hat{P}^O_b - P^O_b ||
\end{equation}
where $J=21$ is the number of hand joints and $B=8$ is the number of object corners 
In the above formula, we omitted all dependencies on the voxels $v$ for brevity.
 
\subsection{VoxelNet: 3D Voxelized Shape}
\label{sec:voxelnet}
Given $V_D$,
 $\hat{P}^H$ and $\hat{P}^O$, the second component of our pipeline, $VoxelNet$, predicts
hand-object voxelized shapes $\hat{V}^H(v)$ and $\hat{V}^O(v)$ for all existing voxels $v$ in the grid.
The 3D CNN-based architecture of $VoxelNet$ is inspired by the work of Malik~\etal~\cite{HandVoxNet2020}. We introduce an additional 3D convolutional layer to predict the voxelized object shape. 
Our voxelized hand-object shapes are defined in the range $[0, 1]$, as done in the previous work. 
The predicted voxelized shapes 
represent complete surface representation for both the hand and the object, including both visible and occluded surface information, as learned from the dataset. 
Thus, entailing richer information for the next algorithmic steps.

To train $VoxelNet$, we use the per-voxel combined sigmoid activation function with binary cross-entropy loss
for the voxelized hand shape (similarly for the voxelized object shape):
\begin{equation}
    \mathcal{L}_{voxel}^H(v) = -(V^H log(\hat{V}^H) + (1 - V^H) log(1 - \hat{V}^H))
\end{equation}
where $V^H$ and $\hat{V}^H$ are the ground truth and the estimated voxelized hand and object shapes, respectively. 
%
%

\subsection{ShapeGraFormer}
\label{sec:graformer}
To obtain topologically-coherent hand-object mesh representation, from the voxelized hand and object predictions obtained from $VoxelNet$, we train a new $ShapeGraFormer$ \cite{Graformer2022}. 
The $ShapeGraFormer$ includes Graph Convolutional layers and Multi-headed Attention layers to convert depth features into a pose and a shape for both the hand and the object. 
Although it was originally designed to lift 2D poses to 3D, GraFormer's design combines the advantages of Graph Convolutional Networks and Transformers making it capable of effectively solving any problem that can be represented as a graph. 
Both the hand and the spherical objects have consistent topology and hence can be represented as a graph.
Therefore, in our method, we utilize three separate GraFormers, namely: hand GraFormer, object GraFormer, and refinement GraFormer, with two feature extractors: a hand feature extractor and an object feature extractor as described in the following subsection. 

Specifically, the network outputs hand-object vertex coordinates $v^H$ and $v^O$, respectively from the MANO hand model $S^H$ and the sphere-based object approximated shape $S^O$.
 The hand and object vertices loss for training is defined using MSE:
\begin{equation}
    \mathcal{L}_{shape} = \frac{\sum_i ( v^H_i - \hat{v}^H_i )^2}{|v^H|} + \frac{ \sum_j ( v^O_j - \hat{v}^O_j )^2}{|v^O|}
\end{equation}
where $\hat{v}^H_i$ and $v^H_i$ are respectively the predicted and ground-truth $i$-th hand vertex coordinates, and $\hat{v}^O_j$ and $v^O_j$ are predicted and ground-truth object vertex coordinates.

\subsubsection{Feature Extractor and Graph Initialization}

For the graph initialization, we use the outputs of the $VoxelNet$ and the raw voxelized depth as shown in Fig. \ref{fig:pipeline}. 
Namely, we extract feature maps $\mathcal{F}_{V}$ from the $VoxelNet$ and pass them through an MLP that converts them into a feature vector of size $256$. 
Additionally, we utilize a simple 3D CNN to extract features from the voxelized shapes and reduce them to $128$ features. 
Furthermore, a 3D Max Pool layer reduces the size of the raw voxelized depth from $44\times44\times44$ to $11\times11\times11$ creating a $1331$ feature vector.
These features are then combined to create a hand feature vector $\mathcal{F}^H_{1715}$ of size $1715$ to initialize the graph vertices as shown in Fig. \ref{fig:pipeline}. The same operation is repeated for the object's vertices resulting in $\mathcal{F}^O_{1715}$.
To initialize the adjacency matrix of the graph layers, we use the mesh faces of the MANO model along with the faces of the spherical mesh as described in Section \ref{sec:model}.



\subsubsection{Positional Embedding}
In order to generate distinct features for each vertex in the graph, we propose a positional embedding layer that converts the vertices of the template meshes for both the hand and object into positional vectors $\mathcal{E}_{i}^{p}$ of the same size as the feature vector where \textit{i} is the index of the \textit{i}th vertex in the combined hand-object graph.
The positional vectors $\mathcal{E}_{i}^{p}$ are then accumulated with the shared feature vectors $\mathcal{F}^H_{1715}$ and $\mathcal{F}^O_{1715}$ depending on whether the \textit{i}th vertex belongs to the hand or the object in order to create a unique representation for each vertex.
For the hand mesh, we adopt the default MANO hand as the template, while for the object mesh, we use the sphere that is utilized for deforming objects, see Section \ref{sec:model}.

\subsubsection{GraFormer Details}
\begin{figure}
\begin{center}
  \includegraphics[width=\linewidth]{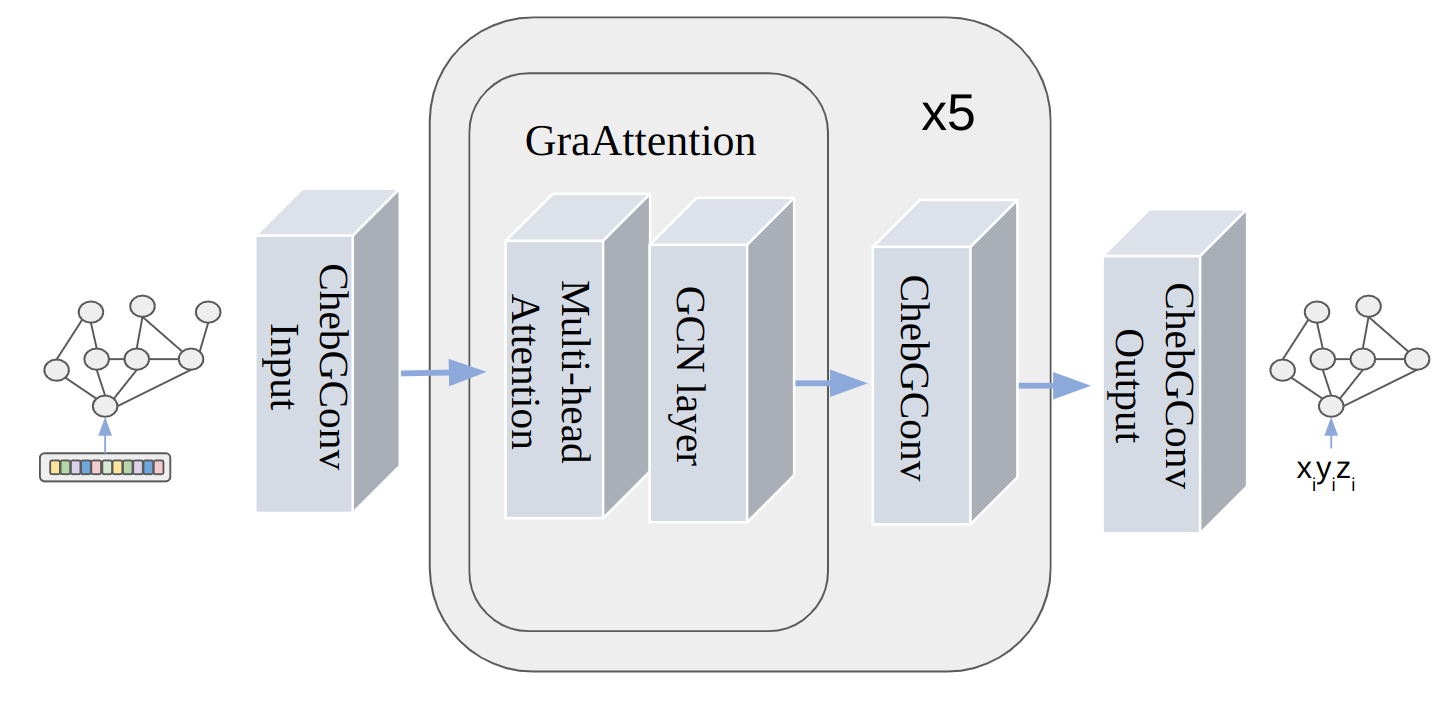}
\end{center}
  \caption{Illustration of the GraFormer architecture.}
\label{fig:graformer}
\end{figure}

Each GraFormer consists of five consecutive GraAttention components followed by Chebyshev Graph Convolutional layers as described in~\cite{Graformer2022} and shown in Fig. \ref{fig:graformer}.
A GraAttention component consists of a Multi-headed Attention layer where the hidden dimension is $128$ and the number of heads is $4$, following the ablation study mentioned in~\cite{Graformer2022} where they studied the impact of hidden dimension size and number of layers in the GraFormer on pose estimation.
Compared to normal Transformers, the last fully connected layer of the Multi-headed Attention in the GraFormer is a graph convolutional layer, not a feed-forward layer. 
The GraFormer also contains an input layer that maps the feature vector to the hidden dimension size and an output layer that maps the hidden dimension into the corresponding 3D coordinate value for each vertex in the graph.

\subsubsection{Refinement GraFormer}
To enhance the realistic appearance and correct minor shape-related artifacts in the predicted object shapes, we add another GraFormer for refinement. We utilize the initial shape produced by the Hand and Object GraFormer as input to another positional embedding layer while we use $\mathcal{F}^H_{1715}$ and $\mathcal{F}^O_{1715}$ to initialize the new graph. 
In the case of using a refinement GraFormer during training, we get $2$ separate shapes from the $ShapeGraFormer$ and apply the same loss function mentioned in \ref{sec:graformer} on both of them.

\subsection{Training and Implementation Details}
\label{sec:training}
We train all the components of our network on fully annotated public hand-object pose and shape datasets, namely, HO-3D~\cite{hampali2020honnotate} and DexYCB~\cite{dexycb}.
For training, we use the Adam~\cite{adam} optimizer with a learning rate set to $.001$. 
For improved convergence, we train $PoseNet$ separately, then fix the obtained weights to train the remaining network components.
All learning and inference are implemented in PyTorch and are conducted on an NVIDIA A100 GPU.


\section{Experiments}
\label{sec:experiments}
In this section, we quantitatively and qualitatively evaluate the effectiveness of our 3D hand-object reconstruction pipeline on two popular datasets, namely HO-3D dataset (version v2 and v3)~\cite{hampali2020honnotate} and DexYCB \cite{dexycb}.
We additionally compare our pipeline with state-of-the-art approaches in Section~\ref{sec:evaluation}. For quantitative evaluation and comparisons,
we use the following two metrics: (i) the average 3D joint location error ($\mathcal{J}$ err.), and (ii) the average 3D vertex location error ($\mathcal{V}$ err.) over all test frames.

\begin{figure*}[p]
\begin{center}
  \includegraphics[width=\linewidth]{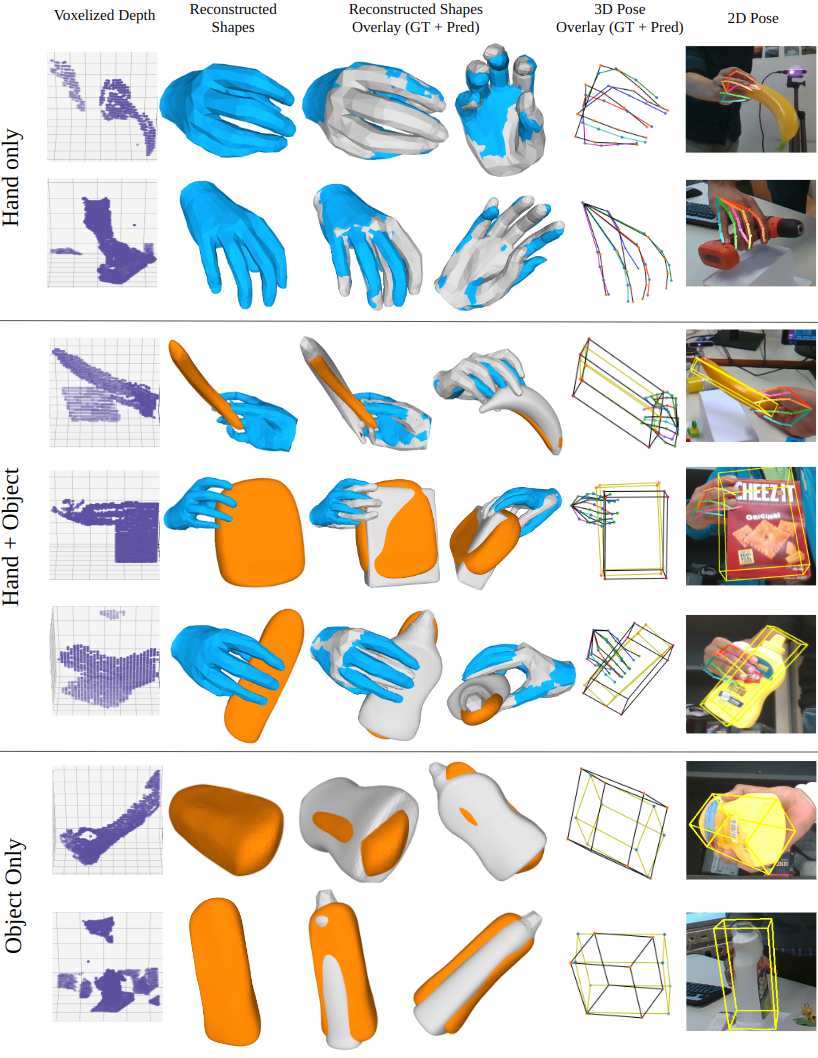}
\end{center}
  \caption{Reconstruction results at different steps. Each row shows a different sample from $\mathcal{D}^{valid}$ or $\mathcal{D}^{eval}$ from HO-3D (v3).} 
\label{fig:ReconstructionEval}
\end{figure*}

%
\begin{figure}
\begin{center}
  \includegraphics[width=\linewidth]{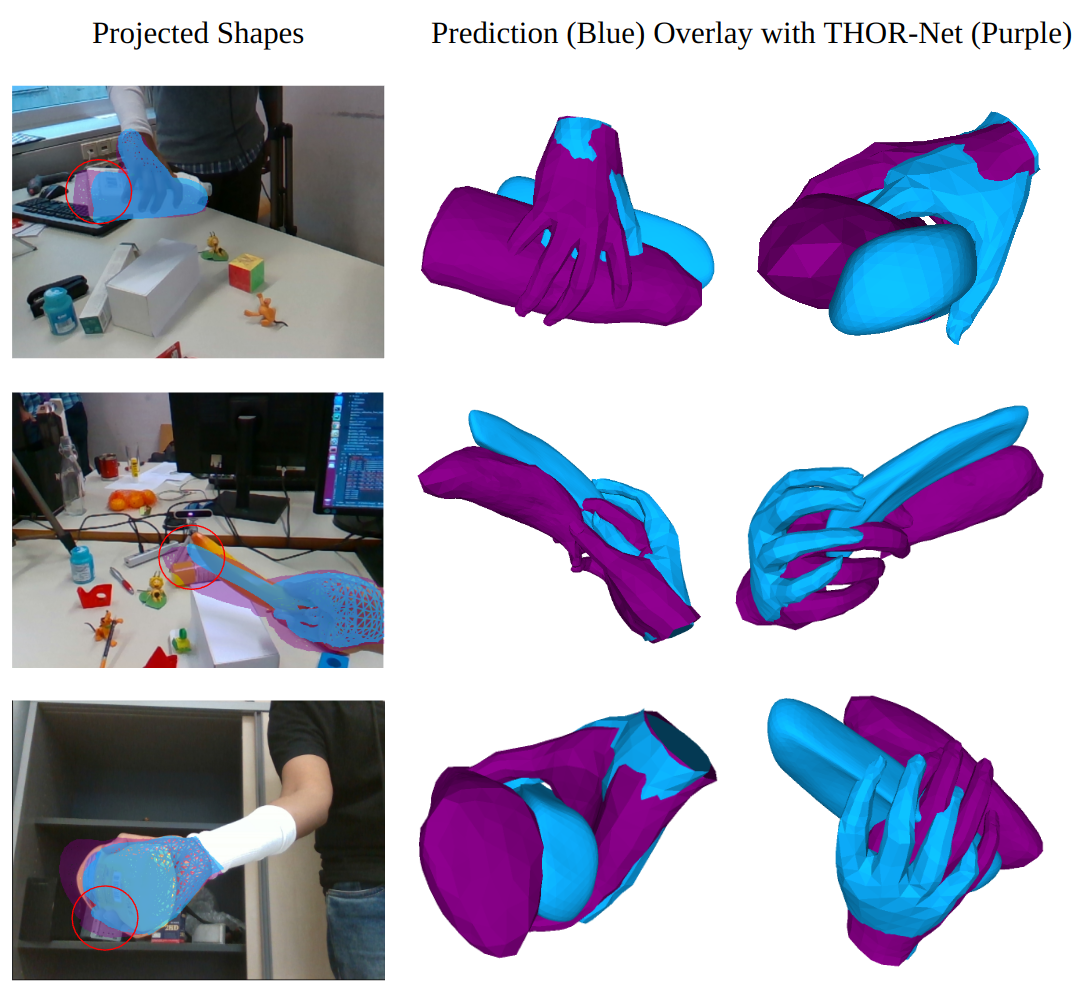}
\end{center}
  \caption{Comparison of our hand-object reconstruction with the THOR-Net pipeline on the evaluation set $\mathcal{D}^{eval}$. Our hand-object reconstruction approach is shown to generate more accurate hand and object shapes.}
\label{fig:THORComparison}
\end{figure}
\begin{figure*}[!t]
\begin{center}
  \includegraphics[width=\linewidth]{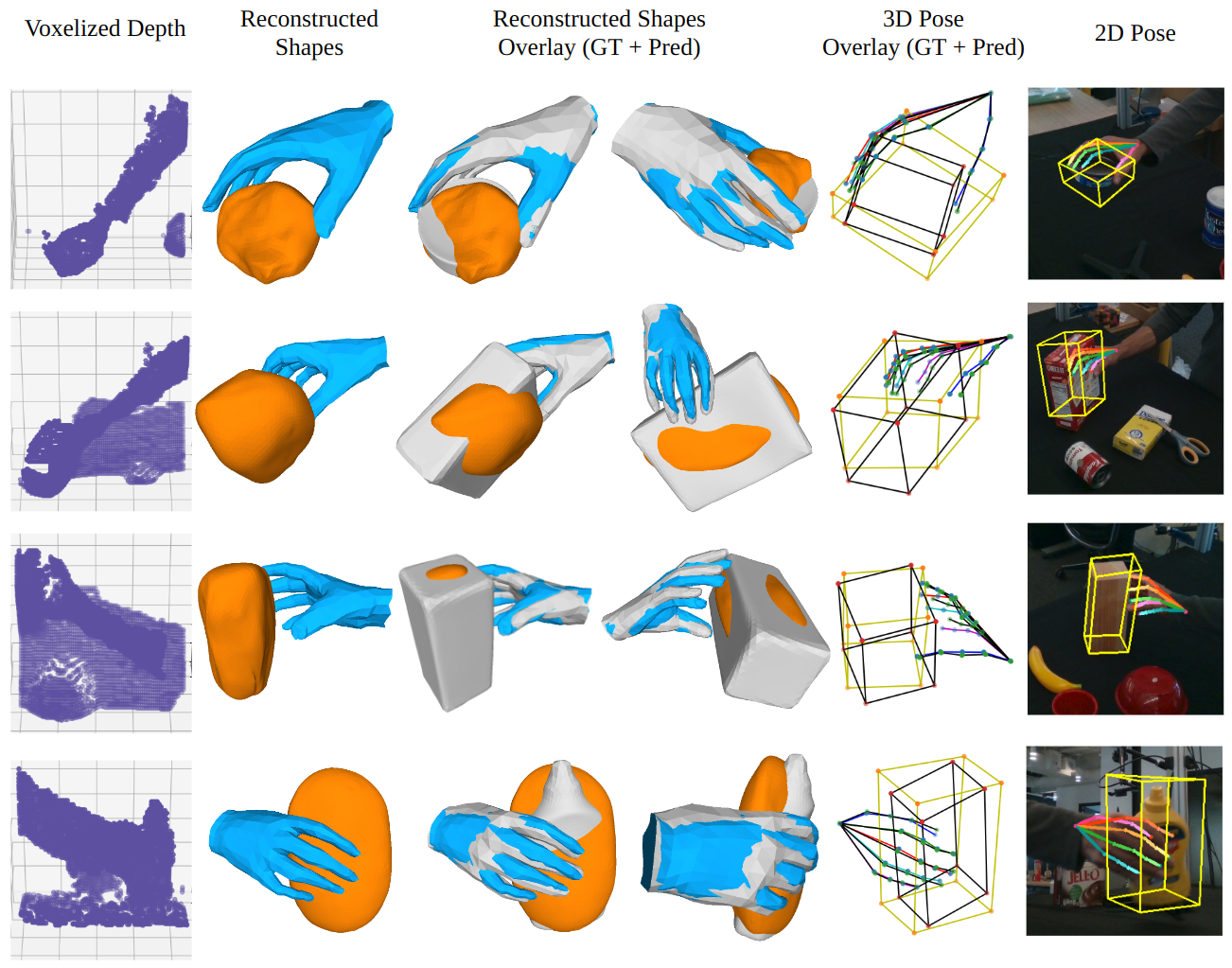}
\end{center}
  \caption{Reconstruction results at different steps. Each row shows a different sample from $\mathcal{D}^{eval}$ from DexYCB. 
  }
\label{fig:dexycb}
\end{figure*}

\subsection{Datasets}
\label{sec:dataset}

\paragraph{HO-3D}
The HO-3D dataset~\cite{hampali2020honnotate}, is a publicly available dataset with 3D pose annotations for hands interacting with objects captured from third-person views. The dataset has multiple versions and we report the results on v2 and v3 of the dataset. 
For HO-3D (v3), the training set $\mathcal{D} := \{\mathcal{D}^{train}, \mathcal{D}^{valid}\}$ contains annotations for $|\mathcal{D}^{train}|=71,662$ and $|\mathcal{D}^{valid}|=10,927$ images and a total of $55$ sequences and $10$ different objects from the YCB-Video dataset~\cite{Xiang2018}, $9$ for the training set and $1$ (unseen) for the evaluation set.
The evaluation set $\mathcal{D}^{eval}$ comprises $13$ sequences with a total of $|\mathcal{D}^{eval}| = 20,137$ frames addressing challenging scenarios: namely, (i) $3$ sequences with ($2$) seen objects and seen hands, (ii) $5$ sequences with ($1$) seen object but unseen hands, and (iii) $5$ sequences with seen hands but $1$ unseen object. Hands in the evaluation set are just annotated with the wrist coordinates, while the full hand is not annotated. Object pose and shape are annotated over all available sets. 

\paragraph{DexYCB}
To extend our evaluation, we also train our network on the DexYCB dataset~\cite{dexycb}. DexYCB contains hand pose and shape and 6D object pose annotations for ~$582k$ frames recorded on $10$ different subjects, using $20$ different objects and $8$ views. We use the S1 evaluation setup as specified by the authors where $\mathcal{D}^{valid}$ contains $1$ unseen subject and $\mathcal{D}^{eval}$ contains $2$ unseen subjects and $\mathcal{D}^{train}$ contains 7 subjects and all different objects. The exact split sizes are: $|\mathcal{D}^{train}|=407,088$, $|\mathcal{D}^{valid}|=58,592$, and $|\mathcal{D}^{eval}|=116,288$.

In the next section, we show results on $\mathcal{D}^{eval}$ for both datasets.

\subsection{Evaluation}
\label{sec:evaluation}

In this section, we evaluate the hand-object shape and pose reconstruction and compare it to state-of-the-art approaches in challenging scenarios. 
%
\paragraph{Methods for Comparison}
\label{sec:evaluation_hand}
%
We compare our work on HO-3D quantitatively with six different methods.
The work of Hasson~\etal~\cite{hasson2019learning} and THOR-Net ~\cite{Aboukhadra_2023_WACV} are the most related to ours in terms of the goals.
However, as they focus on RGB inputs, comparisons are made up to scale. Malik~\etal~\cite{HandVoxNet2020} on the other hand, is based on depth inputs and has a comparable voxel-based network pipeline to our design. However, they ignore the presence of an object and reconstruct hands in isolation. We include the representative RGB-based hand reconstruction approach of Hampali~\etal~\cite{hampali2020honnotate}, and ArtiBoost ~\cite{yang2022artiboost} for completeness.
For a fair comparison, all methods have been (re-)trained on the HO-3D dataset. The first two methods provide publicly available results (hand-only), which we report in Table~\ref{tab:comparison}. We re-implemented HandVoxNet following the authors' instructions and trained all the network components on HO-3D.
In addition to that, we also report the root-relative pose estimation error in Table~\ref{tab:DexYCB} and compare it to two benchmark methods mentioned by the DexYCB authors. 
We also show qualitative samples for hand reconstruction from DexYCB in Fig.~\ref{fig:dexycb}.
%
\paragraph{Hand Reconstruction}
Quantitative evaluation on hand pose and shape reconstruction, shows that our method outperforms existing methods, see Table~\ref{tab:comparison} and Figures~\ref{fig:ReconstructionEval}, \ref{fig:THORComparison}. 
\begin{table}[]
    \centering
    \begin{tabular}{|l|c|c|}
        \hline
        Method & $\mathcal{J}$ err. (\textit{cm}) & $\mathcal{V}$ err. (\textit{cm}) \\
        \hline
        Hampali \etal~\cite{hampali2020honnotate} & 8.42 & 8.34 \\
        Hasson \etal~\cite{hasson2019learning} & 4.97 & 4.96 \\
        HandVoxNet~\cite{HandVoxNet2020} & 2.77 & 2.67 \\ 
        HandVoxNet++~\cite{HandVoxNet++2021} & 2.46 & 2.70\\ 
        THOR-Net \cite{Aboukhadra_2023_WACV} & 2.63 & 2.63 \\
        Ours (hand-object, w/ contact loss) & 2.27 & 2.20
        \\
        \textbf{Ours (hand-only, w/o refinement)} & \textbf{2.13} & \textbf{2.07} \\ 
        \hline
        ArtiBoost \cite{yang2022artiboost} & 2.26 & 10.95 \\
        THOR-Net \cite{Aboukhadra_2023_WACV} & 2.56 & 2.37 \\
        Ours (hand-object, w/ refinement) & 2.15 & 2.09 \\ 
        Ours (hand-only, w/ refinement) & 2.00 & 1.94 \\ 
        \textbf{Ours (hand-only, w/o refinement)} & \textbf{1.99} & \textbf{1.94} \\ 
        \hline
    \end{tabular}
    \caption{Quantitative results on the evaluation set $\mathcal{D}^{eval}$ of HO-3D~\cite{hampali2020honnotate} dataset. 
    Upper table: HO-3D (v2), lower table: HO-3D (v3).
    }
    \label{tab:comparison}
\end{table} 
We achieve a minimum pose and shape prediction improvement respectively of $0.27$ \textit{cm} and $0.43$ \textit{cm} over the state-of-the-art. 
The experiments show that training the network for hand reconstruction alone without object reconstruction improves hand reconstruction and that the refinement stage has no impact on hand reconstruction as shown in Table~\ref{tab:comparison}.
However, joint training for the hand and the object outperforms other methods. 
Fig.~\ref{fig:THORComparison} shows our results on a few frames of the challenging evaluation sequence compared to THOR-Net. In comparison to THOR-Net, our approach reconstructs more accurate hand shapes, as it better exploits hand-object kinematic correlation and depth information. These are implicitly learned while predicting both interacting shapes simultaneously. 
In Table~\ref{tab:DexYCB}, we show an improvement of $0.53$ \textit{cm} in hand pose estimation compared to benchmark results on the DexYCB dataset. 
Fig.~\ref{fig:dexycb} also shows the qualitative reconstruction results on DexYCB without the additional refinement network.

%

\begin{table}[]
    \centering
    \begin{tabular}{|l|c|c|}
    
        \hline
        Method & $\mathcal{J}$ err. (\textit{cm}) & $\mathcal{V}$ err. (\textit{cm}) \\
        \hline
        A2J~\cite{xiong2019a2j} & 2.55 & - \\
        Spurr~\etal~\cite{Spurr2020} & 2.27 & - \\
        \textbf{Ours} & \textbf{1.74} & \textbf{2.65} \\
        \hline
    \end{tabular}
    \caption{Root-relative hand pose estimation results on $\mathcal{D}^{eval}$ of DexYCB~\cite{dexycb} dataset (S1) in comparison to benchmark results.}
    \label{tab:DexYCB}
\end{table} 

\paragraph{Object Reconstruction}
\label{sec:evaluation_object}
We found that topological consistency is the key factor allowing $ShapeGraFormer$ to predict smooth vertices point clouds across all sequences, without the need for additional smoothness constraints, see Fig.~\ref{fig:ReconstructionEval}. After sphere-based registration, all objects share the same topology and number of vertices. 
We believe our choice for the sphere resolution to be a good trade-off between approximation quality and the number of vertices. Sphere-based approximation implicitly repairs irregularities in the target object shapes, making it best suited for hand-object prediction, at the cost of over-smoothed sharp edges and tiny surface details, see Fig.~\ref{fig:objects}.

\begin{figure}
\begin{center}
  \includegraphics[width=\linewidth]{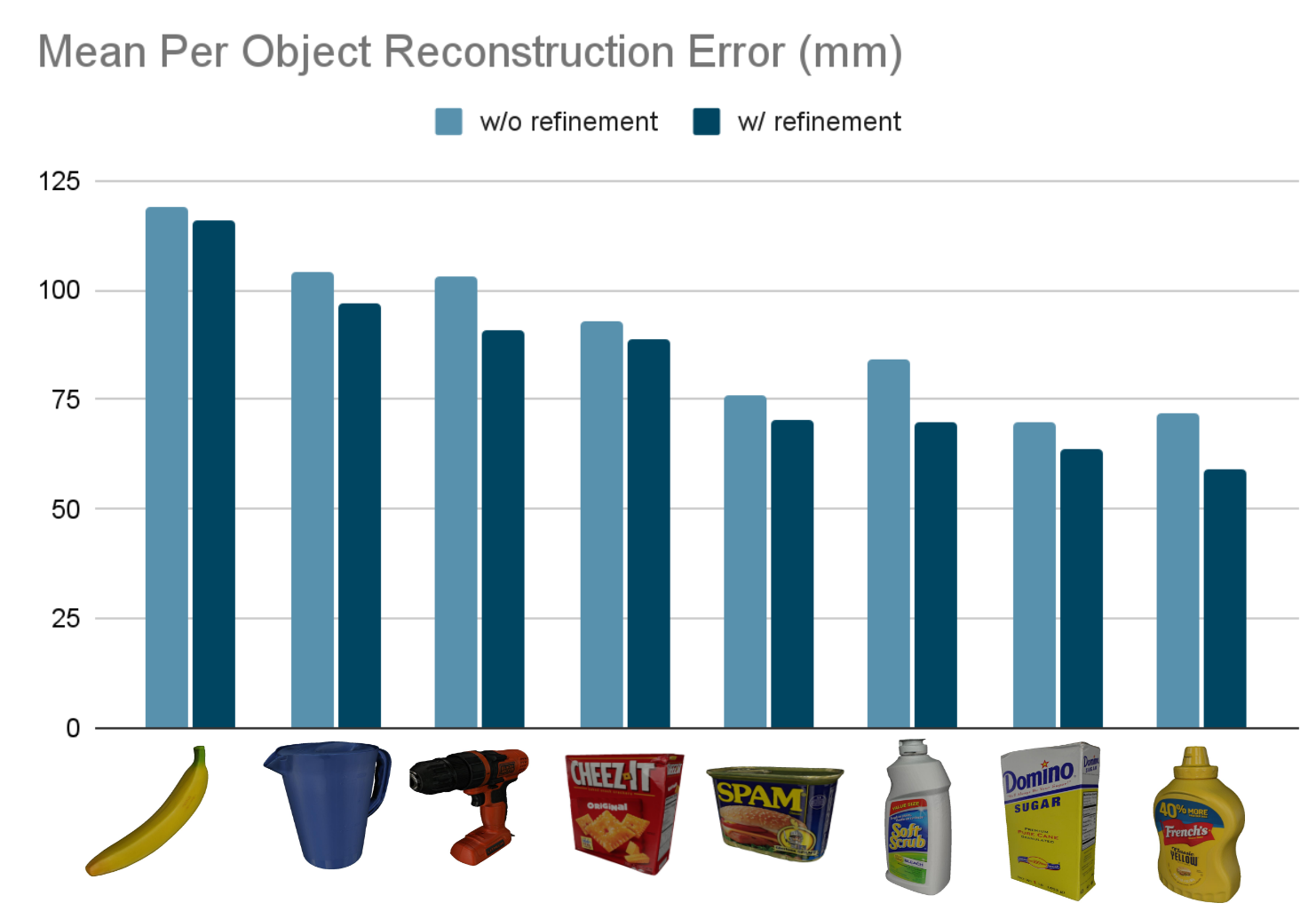}
\end{center}
  \caption{Object reconstruction error ($\mathcal{V}$ err. in \textit{mm}) on different objects from $\mathcal{D}^{valid}$ and $\mathcal{D}^{eval}$ of HO-3D. The results show that adding a refinement GraFormer improves object reconstruction.}
\label{fig:object_evaluation}
\end{figure}

We evaluate object reconstruction on $\mathcal{D}^{valid}$ and $\mathcal{D}^{eval}$ of HO-3D dataset, see Fig.~\ref{fig:object_evaluation} and Table \ref{tab:object_error}, as well as on the DexYCB dataset, see Fig.~\ref{fig:dexycb}.
The additional hand-object refinement step improved object reconstruction error on all objects. This suggests that the refinement network utilizes hand information to improve object reconstruction. 
We notice that even in the presence of inaccurate pose prediction as input, our approach recovers smooth object shapes, see Fig.~\ref{fig:ReconstructionEval}. 
Compared to THOR-Net our approach tends to oversmooth object edges, see Fig.~\ref{fig:THORComparison}, possibly due to the simplified reconstruction approach.
%

\begin{table}[]
    \centering
    \begin{tabular}{|l|c|c|c|c|c|}
        \hline
        & Bottle & Box & Can & Marker & Wood \\
        \hline
        $\mathcal{V}$ err. (\textit{cm}) & 5.9 & 6.4 & 7.0 & 9.7 & 11.5 \\
        \hline
    \end{tabular}
    \caption{Object's reconstruction error ($\mathcal{V}$ err. (\textit{cm})) on a selected set of objects from HO-3D and DexYCB.}
    \label{tab:object_error}
\end{table} 

\subsection{Ablation Study}
To validate our design choices for the $ShapeGraFormer$, we conduct an ablation study to study the modality of $\mathcal{F}_{1715}^{H}$ and the choice of layers within the GraFormer and study the impact on hand reconstruction.
The ablation study in Table \ref{tab:ablation} shows that combining voxelized depth $V_D$, $\mathcal{F}_{V}$, and $\hat{V}^H$ as input to the $ShapeGraFormer$ yields the best results. Furthermore, the mixture of graph layers with transformers in the design of the GraFormer is critical to achieving the best performance. 

\begin{table}[!ht]
    \centering
    \begin{tabular}{|l|c|c|}
        \hline
        Experiment & $\mathcal{J}$ err. (\textit{cm}) & $\mathcal{V}$ err. (\textit{cm}) \\
        \hline
        ShapeGraFormer (w/o $V_D$) & 2.01 & 2.06 \\
        ShapeGraFormer (w/o $\mathcal{F}_{V}$) & 2.05 & 2.01  \\
        ShapeGraFormer (w/o $\hat{V}^H$) & 2.00 & 1.95 \\
        \textbf{ShapeGraFormer} (w/ $V_D \oplus \mathcal{F}_{V} \oplus \hat{V}^H$) & \textbf{1.99} & \textbf{1.94} \\ 
        \hline
        ShapeGraFormer (w/o GCN) & 19.32 & 19.36 \\
        ShapeGraFormer (w/o Transformer) & 2.06 & 2.01 \\
        \textbf{ShapeGraFormer} (w/ GCN + Transformer) & \textbf{1.99} & \textbf{1.94} \\
        \hline
    \end{tabular}
    \caption{Top: An ablation study on the modality of $\mathcal{F}_{1715}^{H}$. Bottom: An ablation study on the GraFormer design choice.}
    \label{tab:ablation}
\end{table} 
\paragraph{Contact Loss: Penetration Avoidance and Contact Enforcement}
We test the impact of differentiable contact loss, consisting of an attraction $\mathcal{L}_{attraction}$ and a repulsion $\mathcal{L}_{repulsion}$ term, similar to Hasson~\etal ~\cite{hasson2019learning}. 
$\mathcal{L}_{attraction}$ is aimed at enforcing hand-object contact, by penalizing the distance between the object and the fingertips, while $\mathcal{L}_{repulsion}$ penalizes mesh interpenetration. 
As shown in Table~\ref{tab:comparison}, the additional loss term does not introduce a tangible increase in performance. Thus, suggesting our independent Refinement GraFormer component implicitly and successfully learns valid hand-object interactions.

%
%

\section{Conclusion}
\label{sec:conclusion}
In this paper, we propose one of the first methods for realistic hand-object pose and shape reconstruction from a single depth map. 
We introduce a novel 3D voxel-based GraFormer network pipeline, which reconstructs detailed 3D shapes via direct regression of mesh vertices. 
We conduct an ablation study to show the effectiveness of our design choices and the impact of utilizing the power of GCNs along with Transformers for hand-object shape estimation and refinement.
We perform quantitative and qualitative analysis on the HO-3D dataset~\cite{hampali2020honnotate} and DexYCB dataset~\cite{dexycb} and show outstanding comparative results with the state-of-the-art.
In future work, we plan to address limitations such as inaccurate annotations. In addition, we will study RGB+D methods to utilize the extra features found in RGB frames.



\bibliographystyle{IEEEtran}  
\bibliography{egbib}


\begin{IEEEbiography}[{\includegraphics[width=1in,height=1.25in,clip,keepaspectratio]{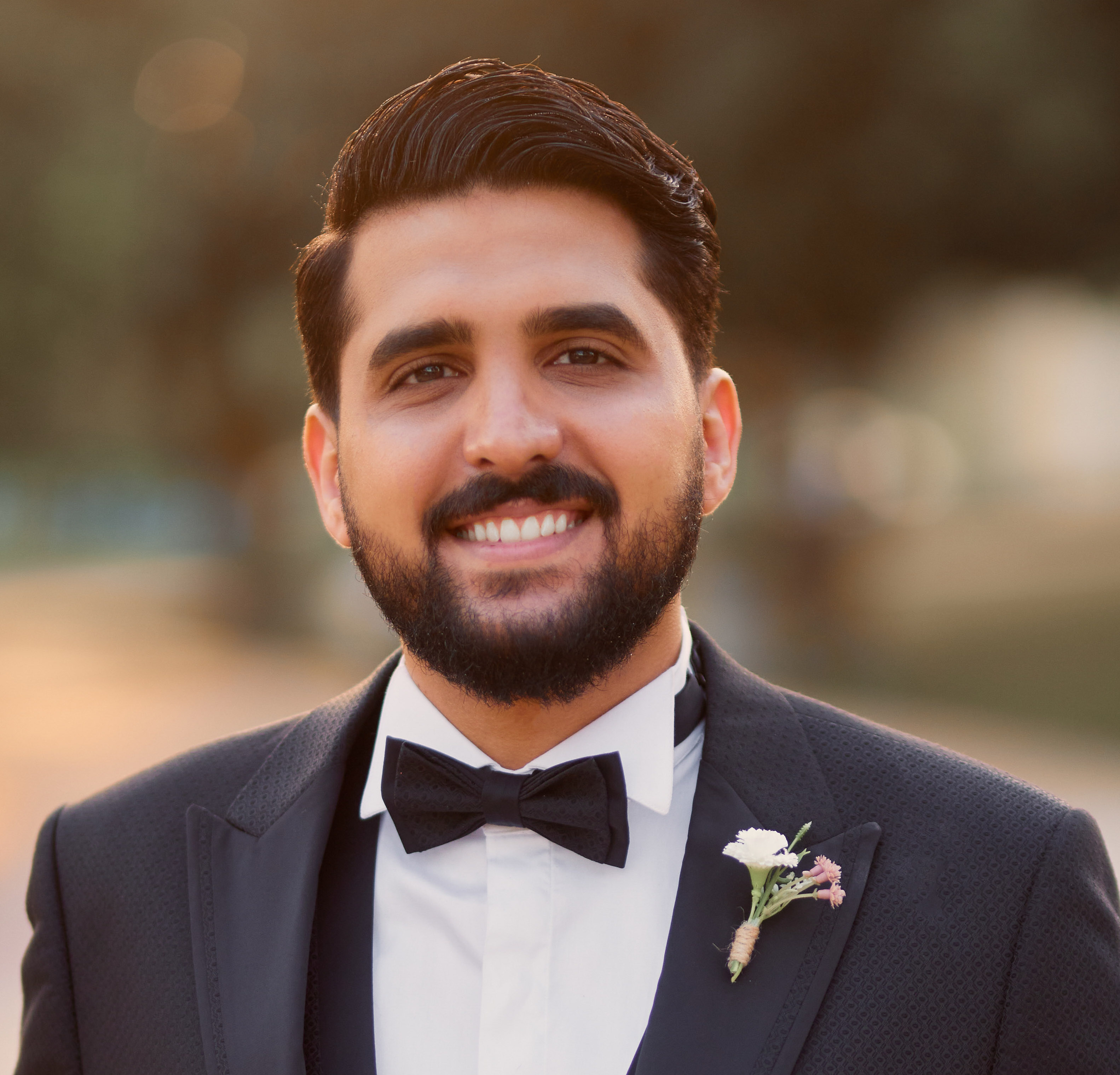}}]{Ahmed Tawfik Aboukhadra} received the B.S. degree in Computer Science from the German University in Cairo (GUC), Egypt, in 2019. 
He received his master's degree in Artificial Intelligence from Maastricht University, the Netherlands, in 2021.
Since 2021, he has been a PhD candidate and a researcher with the Augmented Vision group at DFKI and the Rheinland-Pfälzische Technische Universität Kaiserslautern-Landau (RPTU). His research is focused on Hand-Object 3D reconstruction using Deep Learning. 
\end{IEEEbiography}

\begin{IEEEbiography}[{\includegraphics[width=1in,height=1.25in,clip,keepaspectratio]{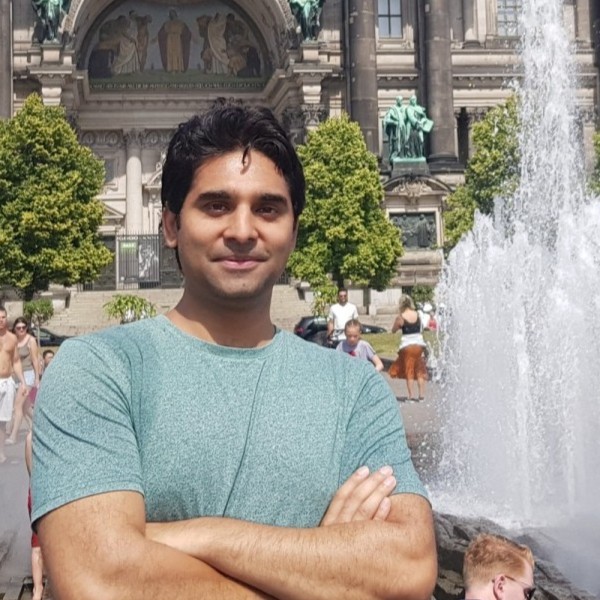}}]{Jameel Malik} received the master’s degree in electrical engineering from the School of Electrical Engineering and Computer Science (SEECS), National University of Sciences and Technology (NUST), Pakistan, and the Ph.D. degree in computer science from Technische Universit at Kaiserslautern, in 2020, for his work on depth-based 3D hand pose and shape estimation. He is an Assistant Professor at NUST-SEECS Pakistan and an adjunct postdoctoral. He is currently a postdoctoral researcher with the Augmented Vision Group, German Research Center for Artificial Intelligence (DFKI GmbH), Kaiserslautern. His current research interests include computer vision, deep learning, and their applications.
\end{IEEEbiography}

\begin{IEEEbiography}[{\includegraphics[width=1in,height=1.25in,clip,keepaspectratio]{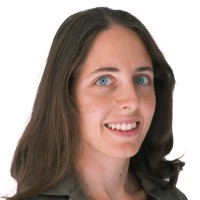}}]{Nadia Robertini} received her master's degree in Visual Computing from Saarland University, Germany, and her Ph.D. degree from the Max Planck Institute for Informatics, Saarbruecken in 2019, for her work on full-body 3D human performance capture. She is currently a Postdoctoral Researcher with the Augmented Vision Group, German Research Center for Artificial Intelligence (DFKI GmbH), Kaiserslautern. Her current research interests include computer vision, machine learning and human performance capture.
\end{IEEEbiography}

\begin{IEEEbiography}[{\includegraphics[width=1in,height=1.25in,clip,keepaspectratio]{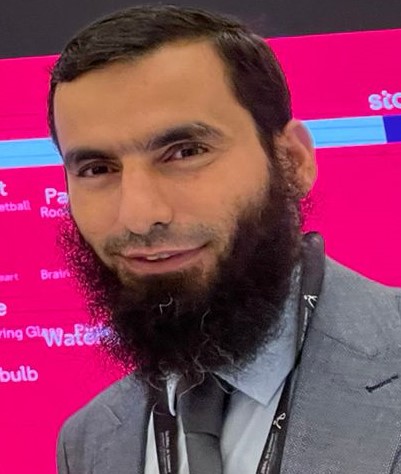}}]{Ahmed Elhayek} received the master’s degree from Saarland University, Germany, in 2010, and the Ph.D. degree from Max-Planck-Institute and Saarland University, in 2015. He worked as a Postdoctoral Researcher with the Augmented Vision Group, German Research Centre for Artificial Intelligence (DFKI). He joined the Faculty of Computer and Cyber Sciences, University of Prince Mugrin (UPM), as an Assistant Professor, in 2018. He founded the Artificial Intelligence department at UPM in 2019. At the moment, he is the head of the Artificial Intelligence department at UPM. 
\end{IEEEbiography}

\begin{IEEEbiography}[{\includegraphics[width=1in,height=1.25in,clip,keepaspectratio]{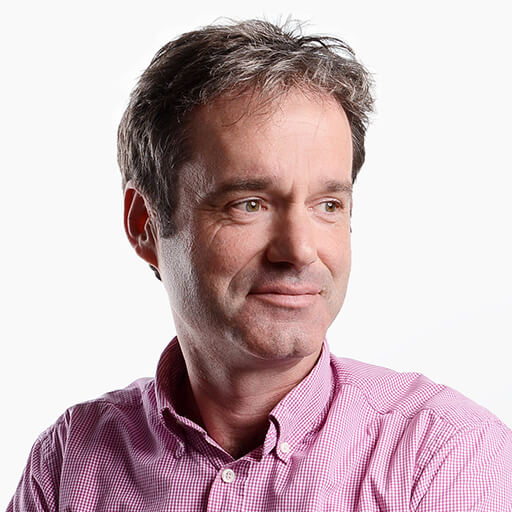}}]{Didier Stricker} 
is a Professor in Computer Science at the University of Kaiserslautern and Scientific Director at DFKI, where he leads the Augmented Vision research group.
He received the Innovation Prize from the German Society of Computer Science in 2006 and got several awards for best papers or demonstrations at different conferences.
He serves as a reviewer for different European or National research organizations. 
He is a reviewer of different journals and conferences in the area of VR/AR and Computer Vision. 
His research interests are cognitive interfaces, user monitoring and on-body-sensor networks, computer vision, video/image analytics, and human-computer interaction.
\end{IEEEbiography}

\newpage





\EOD

\end{document}